\documentclass[10pt,twocolumn,letterpaper]{article}

\usepackage{wacv}
\usepackage{times}
\usepackage{epsfig}
\usepackage{graphicx}
\usepackage{amsmath}
\usepackage{amssymb}

\usepackage{color}
\usepackage{times}
\usepackage{epsfig}
\usepackage{cite}
\usepackage{caption}
\usepackage{subcaption}
\usepackage[linesnumbered,ruled]{algorithm2e}
\usepackage{algpseudocode}
\usepackage{xcolor}
\usepackage{booktabs} 
\usepackage{multirow}
\usepackage{multicol}
\usepackage{balance}
\usepackage{float}
\floatstyle{plaintop}
\restylefloat{table}
\usepackage{array}
\usepackage{algpseudocode}
\usepackage{arydshln}

\DeclareMathOperator*{\argmin}{arg\,min}

\def\wacvPaperID{153} 

\wacvfinalcopy 

\ifwacvfinal
\def\assignedStartPage{1} 
\fi

\ifwacvfinal
\usepackage[breaklinks=true,bookmarks=false]{hyperref}
\else
\usepackage[pagebackref=true,breaklinks=true,colorlinks,bookmarks=false]{hyperref}
\fi

\ifwacvfinal
\else
\fi

\begin{document}
	
	\title{\vspace{-5mm}Exploiting the Redundancy in Convolutional Filters for Parameter Reduction \vspace{-1mm}}
	
	\author{Kumara Kahatapitiya\thanks{work done at University of Moratuwa, Sri Lanka.} \\
		Stony Brook University, Stony Brook, NY\\
		{\tt\small kkahatapitiy@cs.stonybrook.edu}
		\and
		Ranga Rodrigo\\
		University of Moratuwa, Sri Lanka\\
		{\tt\small ranga@uom.lk} 
	}

	\maketitle
	\ifwacvfinal\thispagestyle{empty}\fi

\begin{abstract}
	\vspace*{-3mm} 
	Convolutional Neural Networks (CNNs) have achieved state-of-the-art performance in many computer vision tasks over the years. However, this comes at the cost of heavy computation and memory intensive network designs, suggesting potential improvements in efficiency. Convolutional layers of CNNs partly account for such an inefficiency, as they are known to learn redundant features. In this work, we exploit this redundancy, observing it as the correlation between convolutional filters of a layer, and propose an alternative approach to reproduce it efficiently. The proposed \textit{`LinearConv'} layer learns a set of orthogonal filters, and a set of coefficients that linearly combines them to introduce a controlled redundancy. We introduce a correlation-based regularization loss to achieve such flexibility over redundancy, and control the number of parameters in turn. This is designed as a plug-and-play layer to conveniently replace a conventional convolutional layer, without any additional changes required in the network architecture or the hyperparameter settings. Our experiments verify that LinearConv models achieve a performance on-par with their counterparts, with almost a $50\%$ reduction in parameters on average, and the same computational requirement and speed at inference. Source is available at \url{https://github.com/kumarak93/LinearConv}.
	\vspace*{-5mm}
\end{abstract} 

\section{Introduction}
\label{se:introduction}
Deep Learning has been widely adopted in recent years over feature design and hand-picked feature extraction. This development was supported by the improvement in computational power \cite{jouppi2017datacenter} and large-scale public datasets \cite{deng2009imagenet, caba2015activitynet, cordts2016cityscapes}. In such a resourceful setting, the research community has put forth deep learning models with exceptional performance at the cost of heavy computation and memory usage \cite{simonyan2014very, he2016deep, huang2017densely, szegedy2016rethinking}. Recent studies suggest---although the automated feature learning process captures more meaningful and high-level features---that the learned features of deep learning models inherit a considerable amount of redundancy \cite{denil2013predicting, chakraborty2019feature, song2012sparselet}, which in turn, contributes to unwanted storage  and computations.
Such an inefficiency restricts the deployment of deep learning models in resource-constrained environments. To this end, it is interesting to investigate the possibility of exploiting the redundancy in feature extraction to improve the efficiency of deep networks.

CNNs have become the default backbone of deep neural networks with their success in feature extraction. A set of convolutional layers extracting rich localized features, followed by a set of fully-connected layers linearly combining these extracted features, makes up a general CNN architecture which can achieve state-of-the-art performance in most computer vision tasks. 
However, when we train CNNs, the weights of convolutional layers tend to converge such that there exists a considerable correlation between the learned filters \cite{shang2016understanding, chen2015correlative, wang2017building}. As a consequence, the feature space spanned by these linearly dependent filters is 
a subspace of what could be represented by the corresponding number of filters. Therefore, if carefully optimized, it is possible to span the same feature space with a smaller number of filters, at least theoretically.
Otherwise speaking, the same performance could be achieved with fewer parameters. However, in practice, an over-complete spanning set of filters is allowed, to reach fine-grained performance improvements. Although this is the case, enabling a better control over this redundancy may reveal ways of efficiently replicating the same behavior.

To this end, recent literature has explored the possibility of inducing sparsity \cite{liu2015sparse, song2012sparselet} and separability \cite{chollet2017xception, xie2017aggregated, howard2017mobilenets} in convolutional filters. Although some works consider the inherent correlation in learned convolutional filters for various improvements \cite{shang2016understanding, wang2017building}, it has been overlooked for parameter reduction in deep networks. Moreover, previous works fall short in conveniently controlling the feature redundancy identified as the correlation between features.

\begin{figure*}[t]
	\centering
	\vspace{-4mm}
	\hspace*{-4mm}
	\includegraphics[width=1.1\textwidth]{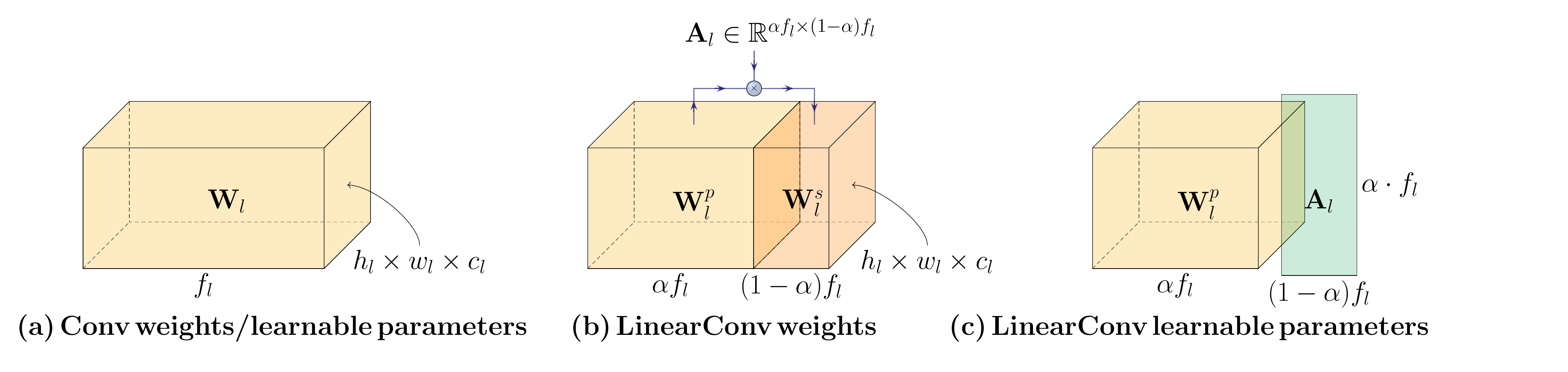} \vspace*{-8mm}
	\caption{
		The weights of Conv (conventional convolution) and the proposed LinearConv layers. The key intuition here is to separate the redundancy and represent it with a smaller number of parameters. Consider a convolutional layer ($l$) with $f_{l}$ number of filters of size $(h_{l}\times w_{l} \times c_{l})$. 
		(a) Conv weights/learnable parameters $\mathbf{W}_{l}$. (b) LinearConv weights: the primary weights $\mathbf{W}^{p}_{l}$ and, the secondary weights $\mathbf{W}^{s}_{l}$ generated by linearly combining the former. The proportion of the primary weights depends on the hyperparameter $\alpha$. (c) Equivalent LinearConv learnable parameters: the primary weights $\mathbf{W}^{p}_{l}$ and the linear coefficients $\mathbf{A}_{l}$. Note that, in LinearConv, we linearly combine the weights of a layer, not the activations. 
	} 
	\vspace*{-4mm}
	\label{Fig:LinearConv}
\end{figure*}
In this paper, we discuss a method to exploit the feature redundancy observed as the correlation between convolutional filters for improving efficiency. The key idea is to restrict the inherent redundancy within the filters, and re-introduce a sufficient amount to maintain the performance, but with a smaller number of parameters.
To do this, we propose a novel convolutional layer \textit{LinearConv} (Fig. \ref{Fig:LinearConv}, \ref{Fig:LinearConvOp}) which consists of two sets of weights: primary weights $\mathbf{W}^{p}_{l}$ , a set of convolutional filters learned with a regularized correlation, and secondary weights $\mathbf{W}^{s}_{l}$, generated by linearly combining the former\footnote{Here, we use the terms \textit{filters} and \textit{weights} interchangeably. However, filters correspond to weights grouped as for each convolution, i.e., weight $\mathbf{W}_{l}$ contains $f_{l}$ filters.}. The linear coefficients $\mathbf{A}_{l}$ are co-learned along with the primary weights. 
The set of secondary weights is spanned by a smaller number of parameters, i.e., linear coefficients, which accounts for the parameter reduction.
In contrast to methods which propose novel or significantly modified architectures for parameter reduction, we design ours to require no change in the architecture or the training process. This allows the LinearConv layer to be plugged into any widely-deployed and tested CNN architectures without any other modifications. We verify that our models can be trained from scratch to achieve a comparable performance, cutting down the number of parameters almost in \textit{half} on average, and can be used in complement with other architectures for network compression.

The main contributions of this paper are as follows:  
\begin{itemize}
	\item We propose a novel convolutional layer LinearConv, which comprises a set of primary and secondary weights. The primary filters are regularized to be linearly independent, whereas the secondary filters are generated by a learnable linear combination of the former. The proposed layer is designed as a plug-and-play component which can replace the conventional convolutional layer without any additional change in the network architecture or the hyperparameter settings. 
	We experimentally validate LinearConv models to achieve a performance on-par with their counterparts, with a reduced number of parameters. 
	\vspace*{1mm}
	\item We introduce a correlation-based regularization loss which gives the flexibility and control over the correlation between convolutional filters. This regularization is applied to the primary filters of LinearConv to make them linearly independent, which in turn, allows a better span for the secondary filters.
\end{itemize}

The rest of the paper is organized as follows: section \ref{se:relatedwork} presents previous works in the area, followed by method and implementation details in section \ref{se:method}, evaluation details in section \ref{se:evaluation}, and finally, conclusion in section \ref{se:conclusion}.  

\section{Related Work}
\label{se:relatedwork}
The capacity of deep neural networks was widely recognized after the proposal of AlexNet \cite{krizhevsky2012imagenet}, which achieved state-of-the-art performance in ILSVRC-2012 \cite{ILSVRC15}. Since then, CNN architectures have improved different facets of deep learning, suggesting deeper \cite{simonyan2014very,he2016deep,he2016identity,huang2017densely}  and wider \cite{szegedy2015going,szegedy2016rethinking,zagoruyko2016wide} architectures. These networks have introduced not only better learning and performance, but also a high demand in computational and storage resources.
\vspace{1mm}

\textbf{Parameter reduction} has been of interest to the community over the years, in parallel to better architectures and optimization techniques. Group convolutions \cite{xie2017aggregated, zhang2018shufflenet,ma2018shufflenet} and depth-wise separable convolutions \cite{chollet2017xception,howard2017mobilenets, sandler2018mobilenetv2} propose to reduce the channel-wise redundancy in learned feature maps.
In \cite{denil2013predicting}, authors predict a majority of the weights using a small subset.
Network pruning techniques either remove non-significant features and fine-tune the network \cite{han2015deep,han2015learning,chakraborty2019feature}, or choose only a set of representative channels \cite{he2017channel}, whereas network quantization methods propose either weight sharing \cite{han2015deep,chen2015compressing} or limited precision of weights \cite{courbariaux2015binaryconnect,courbariaux2016binarized,rastegari2016xnor} to control the memory requirement of deep networks. Furthermore, in \cite{denton2014exploiting, alvarez2016decomposeme, wang2016cnnpack, alvarez2017compression, wen2017coordinating}, authors enforce a low-rank structure or a filter decomposition to reduce network complexity.
More recently, methods such as neural architecture search \cite{zoph2016neural, pham2018efficient, yang2020cars, guo2020hit}, SqueezeNet \cite{iandola2016squeezenet, hu2018squeeze} and EfficientNet \cite{tan2019efficientnet} have improved the state-of-the-art results with reduced complexity. All these works either propose novel network architectures or modify existing architectures. In contrast, we introduce a convenient replacement for convolutional layers, which is applicable to existing network architectures without additional changes.

\vspace{1mm}
\textbf{Correlation in convolutional filters}, and the resulting redundancy have been identified in recent literature. In \cite{shang2016understanding}, authors observe a pair-wise negative correlation in low-level features of CNNs
. A similar property of correlation is identified in \cite{chen2015correlative, wang2017building, chen2018static}, where authors suggest to generate such features based on a separate set of correlation filters instead of learning all the redundant features. Moreover, multiple correlation-based regularization methods have been proposed to de-correlate feature activations \cite{cogswell2015reducing, rodriguez2016regularizing}. From these directions, we can see that the previous works have exploited feature correlation for efficiency.
In this work, we enforce a correlation-based regularization loss on convolutional filters rather than their activations, which restricts them to be a linearly independent set of basis filters.

\vspace{1mm}
\textbf{Linear combinations in convolutional filters} have been explored to improve CNNs in multiple aspects. Separable filters \cite{rigamonti2013learning} and Sparselet models \cite{song2012sparselet} propose the idea of approximating a set of convolutional filters as a linear combination of a smaller set of basis filters. One other direction suggests linearly combining activations instead of the filters which generate them, to efficiently recreate the feature redundancy \cite{jaderberg2014speeding}. In \cite{chen2015correlative, wang2017building}, authors generate correlated filters as a matrix multiplication with a set of correlation matrices: either hand-designed or learned. Here, each primary filter is one-to-one mapped to a dependent filter. We follow a similar process, but instead of learning a one-to-one mapping, we learn a set of linear coefficients which combines a set of learnable filters to generate a set of correlated filters.

Our work is closely related but orthogonal to the recently proposed Octave convolutions \cite{chen2019drop} and Basis filters \cite{li2019learning}, as well as concurrent work LegoNet \cite{yang2019legonet} and GhostNet \cite{han2020ghostnet}. LegoNet \cite{yang2019legonet} assembles a shared set of Lego filters as filter modules, whereas GhostNet \cite{han2020ghostnet} generates new activations based on cheap operations. Octave convolutions \cite{chen2019drop} separate convolutional filters into high-frequency and low-frequency components, processing the latter in low resolution to reduce computations, while keeping the same number of parameters. By design, OctConv is a plug-and-play component similar to ours. However, we follow an intuition based on correlation instead of frequency, reducing the number of parameters. Basis filters \cite{li2019learning} learn a basis of split-weights and linearly combine the activations generated by them, achieving a lower complexity. In contrast, we linearly combine the filters themselves, not their activations. Moreover, Basis filters require the weights of original network for training, which means the \textit{cumulative} computational and memory requirement is higher than the network it replaces. Our method has no such requirement, and can be trained from scratch without additional resources.

In essence, previous works have identified feature correlation and redundancy, utilizing them to improve the efficiency of CNNs. However, such approaches have limited control over the correlation and thus, a narrow outlook on the redundancy and its replication. In contrast, we achieve a finer control over the correlation based on the proposed regularization loss, and a flexible replication of the redundancy in a LinearConv layer. We design our method in a form that can be directly plugged into existing architectures without any additional modifications or hyperparameter tuning, which enables its fast and convenient adoption. 

\section{Method}
\label{se:method}

Our proposition is to control the correlation between convolutional filters in a layer, and in turn, its feature redundancy. To do this, first, we need to restrict the inherent correlation of the convolutional filters that is generally learned in training. Second, we should efficiently introduce a sufficient amount of redundancy to maintain the performance of the network. In this regard, we propose a novel convolutional layer LinearConv, with a primary set of filters which is learned to be linearly independent due to an associated regularization loss, and a secondary set of filters generated through a \textit{learnable} linear combination of the former.

\subsection{Correlation-based Regularization Loss}
\SetKwInput{KwInput}{Input}                
\SetKwInput{KwOutput}{Output}              
\SetKwInput{KwParam}{Parameters} 
\setlength{\textfloatsep}{5pt}
\begin{algorithm}[t]
	\DontPrintSemicolon
	
	\KwInput{Primary weights $\mathbf{W}^{p}_{l=\{1:n\}} \in \mathbb{R}^{\alpha f_{l} \times h_{l} \times w_{l} \times c_{l}}$}
	$L_{c}=0$\;
	\For{$1\leq l \leq n$}    
	{ 
		Reshape $\mathbf{W}^{p}_{l}$ to produce $\mathbf{V}^{p}_{l} \in \mathbb{R}^{\alpha f_{l} \times (h_{l} \cdot w_{l} \cdot c_{l})}$ \;
		Normalize each vector $\mathbf{v}^{p}_{l,i=\{1:\alpha f_{l}\}}$ in $\mathbf{V}^{p}_{l}$, i.e., $\mathbf{v}^{p}_{l,i} \leftarrow \frac{\mathbf{v}^{p}_{l,i}}{\left\lVert \mathbf{v}^{p}_{l,i} \right\rVert_{2}} $\; 
		$L_{c} \leftarrow L_{c} + \left\lVert {\mathbf{V}^{p}_{l}}{\mathbf{V}^{p}_{l}}^\top - \mathbb{I}_{\alpha f_{l}}\right\rVert_{1}$ where $\mathbb{I}_{\alpha f_{l}}$ is the Identity matrix.
	}
	\KwOutput{$L_{c}$}
	\caption{Correlation-based Regularization Loss}
	\label{algo:corrloss}
\end{algorithm}

The intuition of the proposed correlation-based regularization loss is to reduce the inherent redundancy in the primary filters of LinearConv, which we observe as the correlation between the filters. When the correlation is penalized, the primary filters tend to become linearly independent, i.e., orthogonal. This enables the secondary filters to be spanned in a complete subspace which corresponds to the particular number of primary filters (refer subsection \ref{subsec:alpha}). Simply put, the secondary filters get more flexibility to learn the redundant features.

To do this, we define a loss which penalizes the correlation matrix between the flattened primary filters as shown in Algorithm \ref{algo:corrloss}. First, we reshape the input filters $\mathbf{W}^{p}_{l}$ which we wish to make orthogonal, to get a flattened set of filters $\mathbf{V}^{p}_{l}$. Here, we want the correlation matrix to converge to the identity matrix, and hence, define the $\ell1$-norm of the difference between the two to be our regularization loss $L_c$. In contrast to the conventional $\ell1$ regularization, which is applied directly to the weights, our regularization is applied to the difference between the correlation and the identity matrix. At training, $\lambda L_c$ is added to the target loss, where $\lambda$ corresponds to the regularization strength. 
\subsection{LinearConv operation}
Once we regularize the inherent redundancy in the convolutional filters, we need to re-introduce a sufficient redundancy to maintain the performance. We observe a considerable performance drop otherwise. The motivation behind removing and then re-applying the redundancy is to have a control over it. We do this by proposing the LinearConv layer, which has a set of orthogonal primary filters and a set of strictly linearly dependent secondary filters. 

\begin{algorithm}[t]
	\DontPrintSemicolon
	\caption[LinearConv]{LinearConv\footnotemark$\;$Forward Pass}
	\label{algo:LinearConv}
	\KwParam{Primary weights $\mathbf{W}^{p}_{l} \in \mathbb{R}^{\alpha f_{l} \times h_{l} \times w_{l} \times c_{l}}$, Linear coefficients $\mathbf{A}_{l} \in \mathbb{R}^{\alpha f_{l} \times (1-\alpha) f_{l}}$}
	\KwInput{Activations of previous layer $\mathbf{X}_{l-1}$}
	$\tiny{{}^*}$Reshape $\mathbf{W}^{p}_{l}$ to produce $\mathbf{V}^{p}_{l} \in \mathbb{R}^{\alpha f_{l} \times (h_{l} \cdot w_{l} \cdot c_{l})}$ \;
	$\tiny{{}^*}$$\mathbf{U}^{s}_{l} \leftarrow \mathbf{A}_{l}^\top \mathbf{V}^{p}_{l}$ \;
	$\tiny{{}^*}$Reshape $\mathbf{U}^{s}_{l}$ to produce $\mathbf{W}^{s}_{l} \in \mathbb{R}^{(1-\alpha) f_{l} \times h_{l} \times w_{l} \times c_{l}}$ \;
	$\tiny{{}^*}$$\mathbf{W}_{l} \leftarrow [\mathbf{W}^{p}_{l}, \mathbf{W}^{s}_{l}] $ where $[\cdot,\cdot]$ represents concatenation over the first dimension.\;
	$\mathbf{X}_{l} \leftarrow \textnormal{Activation\footnotemark\;}(\mathbf{X}_{l-1} \circledast \mathbf{W}_{l})$
	
	\KwOutput{$\mathbf{X}_{l}$}
\end{algorithm}

\begin{figure}[t]
	\centering
	\hspace*{1mm}
	\includegraphics[width=0.5\textwidth]{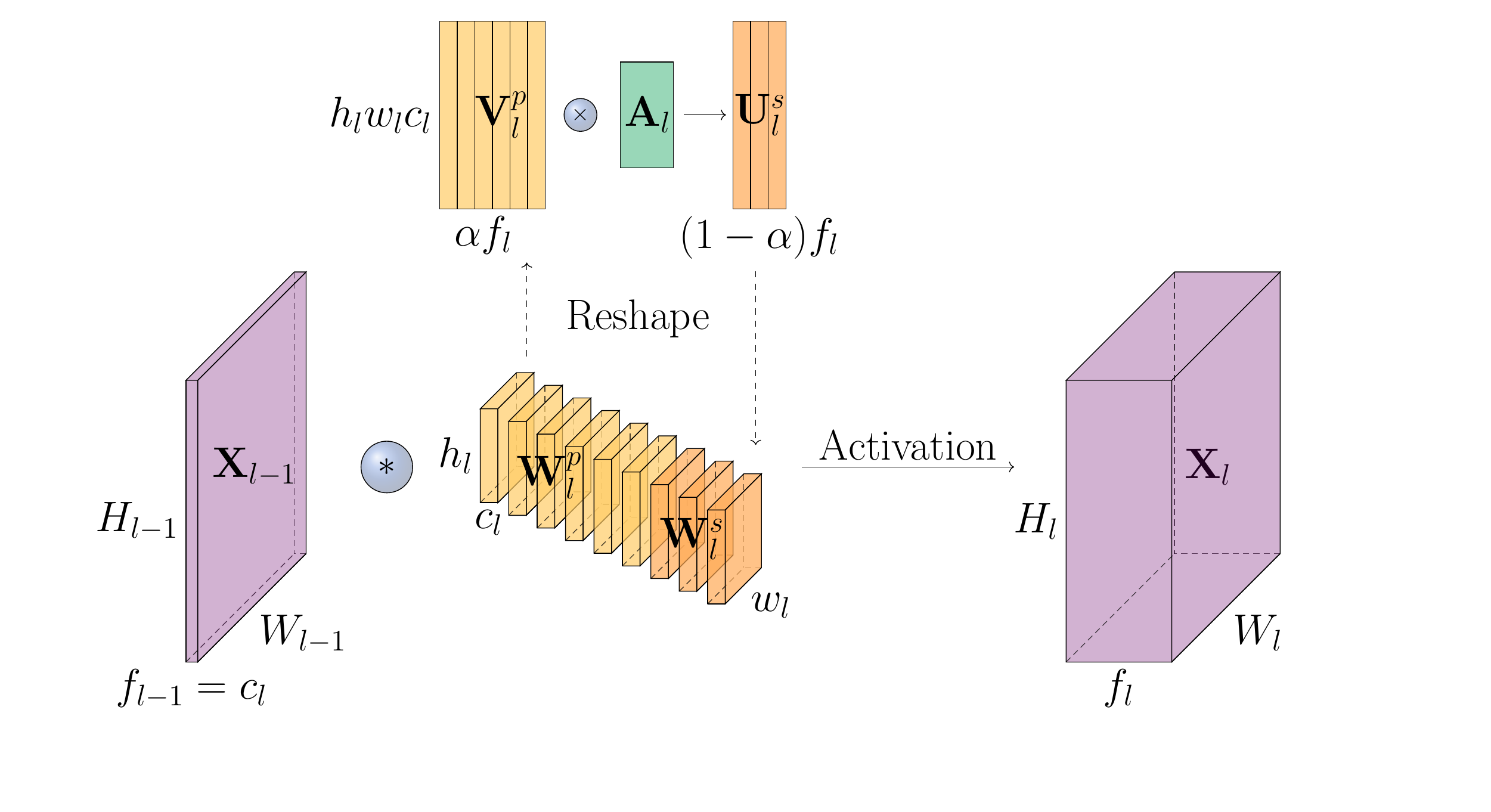} \vspace*{-10mm}
	\caption{LinearConv operation. Each input goes through the same number of operations as in a Conv layer, having no change in either the network architecture or the data flow. The only difference is the weight update process, which is performed in each forward pass \textit{only during training}.
	} 
	\label{Fig:LinearConvOp}
\end{figure}

\footnotetext[2]{Note that the steps marked with $^*$ are performed in each forward pass only during training, to update $\mathbf{A}_{l}$ with backpropagation. At inference, these steps are performed only once at the layer initialization.}
\footnotetext[3]{Activation is shown here inside LinearConv layer to keep the consistency of input-output notation. In implementation, it is a separate component outside LinearConv.} The LinearConv operation is presented in Algorithm \ref{algo:LinearConv}. In the forward pass, it takes in the activations of the previous layer $\mathbf{X}_{l-1}$, and outputs the activations to the next layer $\mathbf{X}_{l}$, similar to Conv operation. The learnable parameters of LinearConv are $\mathbf{W}^{p}_{l}$ and $\mathbf{A}_{l}$, in contrast to $\mathbf{W}_{l}$ in Conv. The secondary filters are generated as a linear combination of the primary filters. To update the parameters $\mathbf{A}_{l}$ during backpropagation, we have to perform this linear combination in every forward pass during training. However, at inference, this is only a \textit{one-time} overhead at initialization. This means that LinearConv will have an increased computational cost during training, but a negligible change at inference. Fig. \ref{Fig:LinearConvOp} shows the LinearConv operation.

It is important to note that in LinearConv, the input $\mathbf{X}_{l-1}$ goes through only a single convolution operation to produce  $\mathbf{X}_{l}$, similar to Conv operation. All other operations are performed on weights. Otherwise speaking, no additional operations are performed on data, having no change in the data flow through the network, compared to the architecture where Conv is replaced by LinearConv. The only difference is how the weights are defined and updated through backpropagation. The benefit of having a design with no change in the architecture is twofold: not only it can be plugged into existing widely-deployed and tested architectures, but also it can function in complement with other network compression methods.

The hyperparameter $\alpha$ defines the proportion of primary filters. The number of learnable parameters in LinearConv changes based on $\alpha$. Thus, it plays an important role in parameter reduction, which we discuss further in the following subsection. In our experiments, we choose  $\alpha$ such that both $\alpha f_{l}$ and $(1-\alpha) f_{l}$ are integers, as they correspond to the number of primary and secondary filters.

We further define a rank-reduced version of LinearConv to reduce the computational cost. Here, the matrix of linear coefficients $\mathbf{A}_{l}$ is decomposed to two low-rank matrices: $\mathbf{A}_{l1} \in \mathbb{R}^{\alpha f_{l} \times r}$ and $\mathbf{A}_{l2} \in \mathbb{R}^{r \times(1-\alpha) f_{l}}$, where $r<\min\{\alpha f_{l},\;(1-\alpha) f_{l}\}$.
\begin{equation}
\label{eq:rank}
\mathbf{A}_{l} =\mathbf{A}^{}_{l1}\mathbf{A}^{}_{l2}\;,\quad
\mathbf{U}^{s}_{l} \leftarrow \mathbf{A}^{\top}_{l2} (\mathbf{A}^{\top}_{l1} \mathbf{V}^{p}_{l})
\end{equation}

\subsection{Effect of hyperparameter $\alpha$}
\label{subsec:alpha}

The hyperparameter $\alpha \in (0,1)$ plays an important role in LinearConv. It is defined as the proportion of primary filters, which directly affects the number of parameters. Moreover, it decides the nature of generated secondary filters as we show in the following formulation. 
\vspace{2mm}

\textbf{On the number of parameters:} Let us express the number of parameters in a Conv layer ($p_l^{\textnormal{C}}$) and an equivalent LinearConv layer ($p_l^{\textnormal{LC}}$) as,
\vspace*{-0mm}
\begin{equation*}
p_l^{\textnormal{C}} = f_{l}h_{l}w_{l}c_{l}\;, \quad
p_l^{\textnormal{LC}} (\alpha) = \alpha f_{l}h_{l}w_{l}c_{l} + \alpha(1-\alpha) f_{l}^2\;,
\end{equation*}
where $p_l^{\textnormal{LC}}$ depends on $\alpha$. To have a parameter reduction in the proposed LinearConv layer, the following should hold:
\vspace*{-0mm}
\begin{equation*}
p_l^{\textnormal{LC}} (\alpha) \leq p_l^{\textnormal{C}} \implies 
\alpha \leq \frac{h_{l}w_{l}c_{l}}{f_{l}}
\end{equation*}

In common CNN architectures, usually $c_{l}(=f_{l-1})/f_{l}$ is either $0.5$ or $1$, except at the input layer. Therefore, even with $1\times 1$ convolutions where $h_{l}=w_{l}=1$, the above condition is satisfied for our choice of $\alpha=0.5$ (discussed below). However, for group convolutions, the effective $c_{l}$ becomes $c_{l}/g_{l}$ where $g_{l}$ is the number of groups, in which case, this condition may not hold (discussed in subsection \ref{subsec:class}).
At the extreme, we would like to maximize the compression, i.e., minimize the number of parameters in a LinearConv layer. This can be formulated as,
\vspace*{-1mm}
\begin{align}
\alpha^{*} 
&=\argmin_{\alpha \in (0,1)}\; \big[\alpha f_{l}h_{l}w_{l}c_{l} + \alpha(1-\alpha) f_{l}^2\big]\;,
\label{eq:argmin_alp}
\end{align} 
and considering the concavity of $p_l^{\textnormal{LC}}(\alpha)$
, $\alpha^{*}$ should be \textit{as small as possible}. 
\vspace{2mm}

\textbf{On the expressiveness of the filters:} For Conv, let us define the filters $\mathbf{W}_{l}$ flattened as $\mathbf{Y}_{l} \in \mathbb{R}^{f_{l} \times h_{l}w_{l}c_{l} }$, and for LinearConv, the primary filters $\mathbf{W}^{p}_{l}$ flattened as $\mathbf{V}^{p}_{l} \in \mathbb{R}^{\alpha f_{l} \times h_{l}w_{l}c_{l} } $, the secondary filters $\mathbf{W}^{s}_{l}$ flattened as $\mathbf{U}^{s}_{l} \in \mathbb{R}^{(1-\alpha) f_{l} \times h_{l}w_{l}c_{l} }$, and the matrix of linear coefficients as $\mathbf{A}_{l} \in \mathbb{R}^{\alpha f_{l} \times(1-\alpha) f_{l}}$. The filters $\mathbf{V}^{p}_{l}$ and $\mathbf{U}^{s}_{l}$ can be represented as,
\vspace*{-0mm}
\begin{equation*} 
\mathbf{V}^{p}_{l} = [\mathbf{v}^{p}_{l,1};\; \cdots;\; \mathbf{v}^{p}_{l,\alpha f}]\;,\quad 
\mathbf{U}^{s}_{l} = [\mathbf{u}^{s}_{l,1};\; \cdots;\; \mathbf{u}^{s}_{l,(1-\alpha) f}]\;, 
\end{equation*}

where $\mathbf{v}^{p}_{l,i} \in \mathbb{R}^{1 \times h_{l}w_{l}c_{l}}$ and $\mathbf{u}^{s}_{l,i} = \sum_{j=1}^{\alpha f_{l}} (\mathbf{A}_{l}^\top)_{i,j} \mathbf{v}^{p}_{l,j} \in \mathbb{R}^{1 \times h_{l}w_{l}c_{l}}$. We see that the span of the filters in a particular layer is important to extract better features, because in subsequent layers, the activations of these filters are re-combined to generate new features. Now, let us consider the vector space spanned by the set of  Conv filters $\mathbf{Y}_{l}$. If we have a sufficient set of linearly independent filters, i.e., if $f_{l}\geq h_{l}w_{l}c_{l}$, we can span a space of $\mathbb{R}^{h_{l}w_{l}c_{l}}$. Otherwise, if $f_{l}< h_{l}w_{l}c_{l}$, the spanned subspace can be re-parameterized as a space of $\mathbb{R}^{f_{l}}$. However, since the Conv filters can be linearly dependent, the spanned subspace can be represented as,
\begin{equation}
\nonumber \text{span}(\mathbf{Y}_{l}) \subseteq \mathbb{R}^{\min\{h_{l}w_{l}c_{l},\;f_{l}\}}.
\end{equation}

In contrast, since the proposed regularization (Algorithm \ref{algo:corrloss}) is applied to the primary filters in LinearConv, $\mathbf{V}^{p}_{l}$ can span a complete space given by,
\begin{equation}
\label{eq:s_v}
\text{span}(\mathbf{V}^{p}_{l}) = \mathbb{R}^{\min\{h_{l}w_{l}c_{l},\;\alpha f_{l}\}}.
\end{equation}

When the primary filters can span a complete space as above, their linear combinations, i.e., the secondary filters have the flexibility to learn any filter in the corresponding space, which in turn results in better activations.
However, the subspace spanned by $\mathbf{U}^{s}_{l}$ further depends on 
the rank of the matrix $\mathbf{A}_{l}$, as we generate $(1-\alpha) f_{l}$ secondary filters based on $\alpha f_{l}$ primary filters.
\setlength{\jot}{3pt}
\begin{align}
\label{eq:s_u}
&\text{span}(\mathbf{U}^{s}_{l})\subseteq
\nonumber\begin{cases}
\text{span}(\mathbf{V}^{p}_{l}) = \mathbb{R}^{\min\{h_{l}w_{l}c_{l},\;\alpha f_{l}\}} & \text{if}\ (1-\alpha) \geq \alpha  \\
\mathbb{R}^{\min\{h_{l}w_{l}c_{l},\;(1-\alpha) f_{l}\}}, & \text{otherwise}
\end{cases}\\
&\text{span}(\mathbf{U}^{s}_{l})\subseteq\mathbb{R}^{\min\{h_{l}w_{l}c_{l},\;\alpha f_{l},\;(1-\alpha) f_{l}\}}
\end{align}

In common CNN architectures, as $f_{l}<h_{l}w_{l}c_{l}$, we can see that the subspace spanned by a LinearConv layer depends on $\alpha$, based on Equations \ref{eq:s_v} and \ref{eq:s_u}. In other words, $\alpha$ acts as a decisive factor for the expressiveness of the feature space. Therefore, we want $\alpha$ to be \textit{as large as possible} to have a good basis set of primary filters. At the same time, we want to generate a \textit{sufficient} redundancy. Considering Equations \ref{eq:argmin_alp}, \ref{eq:s_v} and \ref{eq:s_u}, we can say that $\alpha=0.5$ strikes a balance between the expressiveness of the filters and the number of parameters. In this case, the subspace spanned by the filters in a LinearConv layer becomes,
\vspace*{-2mm}
\begin{equation}
\nonumber\text{span}(\mathbf{U}^{s}_{l})\subseteq\text{span}(\mathbf{V}^{p}_{l})=\mathbb{R}^{\min\{h_{l}w_{l}c_{l},\;\frac{f_{l}}{2}\}},
\end{equation}
and the corresponding number of learnable parameters becomes,
\begin{equation}
\nonumber
p_l^{\textnormal{LC}} = \frac{1}{2}f_{l}\bigg(h_{l}w_{l}c_{l} + \frac{1}{2}f_{l}\bigg).
\end{equation}

\section{Evaluation}
\label{se:evaluation}
\subsection{Implementation details}
\vspace*{-1mm}
\setlength\extrarowheight{2pt}
\newcolumntype{C}[1]{>{\centering\arraybackslash}p{#1}}

We want to evaluate the performance of LinearConv in common CNN architectures. Therefore, we replace the Conv layers in a variety of architectures with LinearConv (details in Appendix A). Except for a baseline CNN, all the others are well-known models proposed in the literature \cite{simonyan2014very, springenberg2014striving, he2016deep, xie2017aggregated, sandler2018mobilenetv2}. AllConv \cite{springenberg2014striving} is a fully convolutional network. The rest of the models have a single fully-connected (\textit{fc}) layer each, which outputs logits. 
ResNeXt-29 is based on group convolutions \cite{krizhevsky2012imagenet, xie2017aggregated}, whereas in MobileNetV2 \cite{sandler2018mobilenetv2}, it becomes depth-wise separable convolutions \cite{chollet2017xception} as the number of groups is equal to the number of input feature maps. In all configurations, convolution is followed by batch normalization \cite{ioffe2015batch} and ReLU activations \cite{nair2010rectified}, except for AllConv, which omits batch normalization. As for pooling, Base and VGG11 use max-pooling, whereas the others use strided convolutions with stride of 2. We carefully selected these specific models to show the behavior of LinearConv in vanilla CNNs, and the networks with skip-connections and group convolutions. All our models are implemented on PyTorch \cite{paszke2017automatic}.
We evaluate our models in multiple classification datasets: CIFAR-10,100 \cite{krizhevsky2009learning}, Street View House Numbers (SVHN) \cite{netzer2011reading}, MNIST \cite{lecun1998gradient} and Fashion MNIST \cite{xiao2017fashion}. All datasets are channel-wise normalized at the input. The images in MNIST and Fashion MNIST are zero padded and randomly cropped. In CIFAR-10,100 and SVHN we perform random horizontal flips and random crops. All inputs are of $32\times32$ resolution.
We trained the models for 250 epochs with Adam optimizer \cite{kingma2014adam} and cross-entropy loss. The proposed regularization loss is added to the target loss, with a regularization  strength of $\lambda =10^{-2}$. 
We use an input batch size of 64 in all experiments. The initial learning rate is selected to be one of $\left\lbrace10^{-3},\; 10^{-4}\right\rbrace$ by observation, with a step-wise decay of $10^{-1}$ once every 100 epochs. 
For each configuration, we report the maximum accuracy on test data, model size (number of learnable parameters) and the computational requirement.

\subsection{Ablation study}
\label{subse:ablation}

In our ablation studies, we investigate different aspects of LinearConv. Specifically, we verify how the hyperparameter $\alpha$ affects the performance, which we discussed previously, and the significance of generating redundant features in the proposed layer.

The hyperparameter $\alpha$ controls the number of primary filters in a LinearConv layer. 
For smaller  $\alpha$, LinearConv will have a larger proportion of secondary filters, extracting more redundant features.
When $\alpha=1$, LinearConv without regularization is equivalent to Conv. Table \ref{Table:abl_alpha} presents the change in performance with $\alpha$. We see a stable accuracy when $\alpha=0.5$, which verifies our claim in subsection \ref{subsec:alpha}. This setting allows a better flexibility for the spanned redundant filters, striking a balance between parameter reduction and performance degradation. In terms of the computational requirement at training, we see a peak in the same setting as well. This is due to the increased dimensions of the matrix of linear coefficients $\mathbf{A}_{l}$. As this is undesirable, we propose a rank-reduced version of the matrix in the following subsection. 
\newcolumntype{P}[1]{>{\raggedleft\arraybackslash}p{#1}}
\begin{table}[t]
	\small
	\begin{center}
		\resizebox{0.48\textwidth}{!}{
			\begin{tabular}{l P{8mm} P{8mm}P{8mm}P{8mm}P{8mm}P{8mm}}			
				\hline\vspace*{0mm}
				$\alpha$ ratio& $0.125$  & $0.25$ & $0.5$ & $0.75$ & $0.875$ & $1$ \\ \hline
				\vspace*{-0.mm}Accuracy (\%) & $-1.2$ & $-0.1$ & $\mathbf{+0.0}$ & $-0.1$ & $-0.3$ & $90.4$ \\ 
				\vspace*{-0.mm}Parameters (M) & $1.30$& $2.54$ & $4.92$ & $7.15$ & $8.21$ & $9.23$ \\ 
				FLOPs (B) & $1.139$& $1.838$ & $2.398$ & $1.842$ & $1.146$  & $0.171$\\
				\hline
			\end{tabular}
		}\vspace{3mm}
		\caption{ Performance of the regularized LinearConv model VGG11-Lr on CIFAR-10 \cite{krizhevsky2009learning} for different $\alpha$. The accuracies are given w.r.t. Conv model VGG11\cite{simonyan2014very}, which corresponds to $\alpha=1$ (not regularized). We present FLOPs for training. At inference, all configurations have the same computational cost as $\alpha=1$. This verifies our claim in subsection \ref{subsec:alpha}, that $\alpha=0.5$ is a reasonable selection for the hyperparameter. Since the increment in FLOPs is proportional to the size of the matrix $\mathbf{A}_{l}$, i.e., $\alpha(1-\alpha) f_{l}^2$, we see a maximum at this setting, which can be remedied with the rank-reduced version of LinearConv as we discuss in the following subsection.
		}
		\label{Table:abl_alpha}
		\vspace{-1mm}
	\end{center}
\end{table}

\begin{table}[t]
	\small
	\begin{center}
		\resizebox{0.48\textwidth}{!}{
			\begin{tabular}{l P{14mm}P{15mm}P{15mm}P{18mm} }			
				\hline
				Model                               & CIFAR-10 (\%) & CIFAR-100 (\%) & Parameters (M) & FLOPs (B) \\ \hline
				\vspace*{-0mm}Base                                 & $87.2$ & $61.0$ & $0.40$ & $0.017$ \\ 
				\vspace*{-0mm}Base-r  & \footnotesize-0.9  & \footnotesize-1.0 & $\mathbf{|\;\;\;}$ & $\mathbf{|\;\;\;}$ \\ 
				Base-e  & \footnotesize-1.0  & \footnotesize-1.9 & $0.22$& $0.010$ \\ \hdashline 
				\vspace*{-0mm}Base-L   & \footnotesize-0.2  & \footnotesize-1.0 & $0.23$ & $(0.017)\;0.060$ \\ 
				\textbf{Base-Lr }  & \footnotesize\textbf{-0.1}  & \footnotesize\textbf{-0.1} & $\mathbf{|\;\;\;}$ & $\mathbf{|\;\;\;}$ \\ \hline
				\vspace*{-0mm}VGG11 \cite{simonyan2014very}                              & $90.4$& $65.4$ & $9.23$ & $0.171$ \\ 
				\vspace*{-0mm}VGG11-r  & \footnotesize-1.1  & \footnotesize-0.7 & $\mathbf{|\;\;\;}$ & $\mathbf{|\;\;\;}$ \\
				VGG11-e  & \footnotesize-1.5  & \footnotesize-1.1 & $4.93$ & $0.093$ \\ \hdashline  
				\vspace*{-0mm}VGG11-L   & \footnotesize-0.2  & \footnotesize-1.0 & $4.92$ & $(0.171)\;2.398$ \\ 
				\textbf{VGG11-Lr}   & \footnotesize\textbf{+0.0}  & \footnotesize\textbf{+0.0} & $\mathbf{|\;\;\;}$ & $\mathbf{|\;\;\;}$ \\ 
				\hline
			\end{tabular}
		}\vspace{3mm}
		\caption{
			Comparison of different versions of Conv and LinearConv models: -r (regularized), -e (equiv-parameter) and -L (LinearConv) (Details in subsection \ref{subse:ablation}). The accuracies are given w.r.t. corresponding Conv model. The values same as above are indicated by $|$, and FLOPs correspond to (inference) training. The degraded accuracy in the regularized Conv model shows the importance of the redundant features. Similarly, the version with an equivalent number of parameters as LinearConv is not able to achieve the same performance. The regularized version of LinearConv show a better performance stability in comparison.
		}
		\label{Table:abl_reg}
		\vspace*{-1mm}
	\end{center}
\end{table}

We want to evaluate whether the proposed LinearConv layer and regularization can provide a performance stability in parameter reduction. To do this, we compare different versions of Conv and LinearConv models. We regularize Conv layers without regenerating any redundancy (-r) to see the importance of secondary filters in LinearConv. We consider Conv models with an equivalent amount of parameters as LinearConv (-e) to understand if the proposed layer is actually required in reducing parameters. To evaluate if the regularization provides any advantage in LinearConv, we consider un-regularized (-L) and regularized (-Lr) versions of LinearConv. Table \ref{Table:abl_reg} shows the performance of these models. The regularized and equiv-parameter versions of Conv models verify that we need redundant features to maintain a stable performance, and simply reducing the number of parameters would not help. Among LinearConv models, the regularized version shows a better stability, verifying our claim in subsection \ref{subsec:alpha} that it provides a better flexibility in spanning redundant filters, and feature subspace in turn.
\subsection{Classification Results}
\label{subsec:class}

\begin{table*}[t!]
	\small
	\begin{center}
		\resizebox{0.98\textwidth}{!}{
			\begin{tabular}{l P{15mm}P{15mm}P{15mm}P{15mm}P{17mm}P{20mm}P{20mm} }			
				\hline
				\vspace*{-1mm}Model & CIFAR-10 & CIFAR-100& SVHN & MNIST & Fashion & Parameters (M)& FLOPs (B)\\
				& (\%)  &  (\%) &  (\%) &  (\%) & MNIST (\%)&  & \\ \hline 
				\vspace*{-0mm}Base                      & $87.2$ & $61.0$ & $92.0$ & $99.3$ & $93.5$ & $0.40$ & $0.017$ \\ 
				\vspace*{-0mm}Base-Lr     & \footnotesize-0.1 & \footnotesize-0.1 & \footnotesize+0.6 & \footnotesize-0.1 & \footnotesize-0.2 & $\scriptstyle(\times 0.43 \downarrow)\;$$0.23$ & $\scriptstyle{(\times 2.5 \uparrow)}$$\;0.060$ \\ 
				Base-LRr & \footnotesize-0.6 & \footnotesize-0.6 & \footnotesize+0.2 & \footnotesize-0.2 & \footnotesize-0.8 & $\scriptstyle\mathbf{(\times 0.48\downarrow)}\;$$0.21$  & $\scriptstyle\mathbf{(\times 0.5 \uparrow)}$$\;0.025 $  \\ \hline 
				
				\vspace*{-0mm}VGG11 \cite{simonyan2014very}                    & $90.4$& $65.4$ & $95.4$ & $99.3$ & $93.8$ & $9.23$ & $0.171$ \\ 
				\vspace*{-0mm}VGG11-Lr     & \footnotesize+0.0 & \footnotesize+0.0 & \footnotesize+0.1 & \footnotesize+0.1 & \footnotesize-0.2 & $\scriptstyle(\times 0.47\downarrow)\;$$4.92$ & $\scriptstyle(\times 13 \uparrow)$$\;2.398$ \\ 
				VGG11-LRr  & \footnotesize-0.6 & \footnotesize-1.1 & \footnotesize+0.0 & \footnotesize+0.1 & \footnotesize-0.2 & $\scriptstyle\mathbf{(\times 0.50\downarrow)}\;$$4.65$ & $\scriptstyle\mathbf{(\times 1 \uparrow)}$$\;0.350$ \\ \hline 
				
				\vspace*{-0mm}AllConv  \cite{springenberg2014striving}                    & $85.0$ & $42.7$ & $94.5$ & $99.0$ & $92.6$ & $1.37$ & $0.315$ \\ 
				\vspace*{-0mm}AllConv-Lr  & \footnotesize-0.9  & \footnotesize-0.2  & \footnotesize+0.1  & \footnotesize-0.1  & \footnotesize+0.1  & $\scriptstyle(\times 0.46\downarrow)\;$$0.74$ & $\scriptstyle(\times 0.4 \uparrow)$$\;0.440$ \\ 
				AllConv-LRr  & \footnotesize\textcolor{gray}{-3.1}  & \footnotesize-1.6  & \footnotesize-1.9  & \footnotesize-0.2  & \footnotesize-0.4 & $\scriptstyle\mathbf{(\times 0.49\downarrow)}\;$$0.70$ & $\scriptstyle\mathbf{(\times 0.1 \uparrow)}$$\;0.344$ \\ \hline 
				
				\vspace*{-0mm}ResNet-18 \cite{he2016deep}                      & $91.9$ & $66.2$ & $96.2$ & $99.4$ & $94.6$ & $11.17$ & $0.558$ \\ 
				\vspace*{-0mm}ResNet-18-Lr   & \footnotesize-0.8  & \footnotesize+2.6  & \footnotesize+0.1  & \footnotesize+0.0  & \footnotesize-0.2 & $\scriptstyle(\times 0.46\downarrow)\;$$6.03$ & $\scriptstyle(\times 4.5 \uparrow)$$\;3.074$ \\ 
				ResNet-18-LRr   & \footnotesize-1.9  & \footnotesize+1.5  & \footnotesize-0.2  & \footnotesize-0.1  & \footnotesize-0.2 & $\scriptstyle\mathbf{(\times 0.50\downarrow)}\;$$5.64$ & $\scriptstyle\mathbf{(\times 0.4 \uparrow)}$$\;0.775$ \\ \hline 
				
				\vspace*{-0mm}ResNeXt-29  \cite{xie2017aggregated}                    & $92.9$ & $76.3$ & $96.2$ & $99.3$ & $94.5$ & $9.13$ & $1.424$ \\ 
				\vspace*{-0mm}ResNeXt-29-Lr   & \footnotesize+0.5  & \footnotesize\textcolor{gray}{-2.1}  & \footnotesize-0.3  & \footnotesize+0.0  & \footnotesize-0.1 & $\scriptstyle(\times 0.29\downarrow)\;$$6.48$ & $\scriptstyle(\times 2.6 \uparrow)$$\;5.089$ \\ 
				ResNeXt-29-LRr  & \footnotesize+0.8  & \footnotesize\textcolor{gray}{-2.9}  & \footnotesize-0.7  & \footnotesize+0.1  & \footnotesize-0.3 & $\scriptstyle\mathbf{(\times 0.48\downarrow)}\;$$4.71$ & $\scriptstyle\mathbf{(\times 0.2 \uparrow)}$$\;1.691$ \\ \hline  
				
				\vspace*{-0mm}MobileNetV2 \cite{sandler2018mobilenetv2}                     & $93.1$ & $73.5$ & $96.1$ & $99.5$ & $93.5$  & $2.30$ & $0.098$ \\ 
				\vspace*{-0mm}MobileNetV2-Lr  & \footnotesize-0.3  & \footnotesize-1.8  & \footnotesize-0.1  & \footnotesize-0.1  & \footnotesize+0.2 & $\scriptstyle\textcolor{blue}{(\times 0.70 \uparrow)}\;$$3.92$ & $\scriptstyle\textcolor{blue}{(\times 166 \uparrow)}$$\;16.398$ \\
				MobileNetV2-LRr  & \footnotesize-1.7  & \footnotesize\textcolor{gray}{-3.0}  & \footnotesize-0.1  & \footnotesize-0.1  & \footnotesize+0.4 &  $\scriptstyle\mathbf{(\times 0.41\downarrow)}\;$$1.35$ &  $\scriptstyle\mathbf{(\times 8.5 \uparrow)}$$\;0.931$ \\ \hline
				
			\end{tabular}
		}\vspace{3mm}
		\caption{Comparison of selected common CNN architectures where the Conv layers are replaced by LinearConv layers. Here, we present the accuracy w.r.t. corresponding baselines, the number of learnable parameters and the computational requirement for training (and relative change). For fair comparison, all the results presented here are for our experiment settings. LinearConv models perform on-par with the respective baselines with a reduced number of parameters. The regularized, rank-reduced LinearConv models (-LRr) show almost a $50\%$ reduction in parameters on average, with a maximum of $\times1$ increment in FLOPs at training, except for the case of MobileNetV2 with depth-wise separable convolutions (discussed in subsection \ref{subsec:class}). A few cases with a significant accuracy drop ($>2\%$) is shown in gray.
		}
		\label{Table:results}
		\vspace*{-5mm}
	\end{center}
\end{table*}

Here, we compare a variety of CNN architectures which highlight different aspects of network design, replacing their Conv layers with the proposed LinearConv layers. For fair comparison, we evaluate all the configurations in our experiment settings, and present the maximum accuracies achieved in Table \ref{Table:results}. The LinearConv models show a comparable performance with the respective baselines with a reduced number of parameters. Specifically, the regularized LinearConv (-Lr) models achieve an average of $44\%$ and a minimum of $29\%$ reduction in parameters (except in MobileNetV2) with performance variations ranging from $-2.1\%$ to $+2.6\%$. However, in terms of the computational requirement, this version shows a maximum increment of $\times13$ in VGG11-Lr, which is undesirable. To remedy this, we introduce a rank-reduced, regularized version of LinearConv (-LRr) as described in Eq. \ref{eq:rank}. We set the rank of the matrix $\mathbf{A}_{l}$ to be 10 for all layers in every configuration by observation. These rank-reduced versions achieve an average of almost $50\%$ and  a minimum of $41\%$  reduction in parameters with performance variations ranging from $-3.1\%$ to $+1.5\%$. The increment in computational requirement for training is contained to a maximum of $\times1$ since the matrix multiplication which generates the secondary filters is decomposed to be lightweight. Under this setting, vanilla CNNs, and CNNs with skip connections, group convolutions and even depth-wise separable convolutions show a desirable parameter reduction with a comparable performance, when LinearConv is introduced. At inference, LinearConv models have the same computational requirement as the respective Conv models, since the extra cost will be only a one-time cost at initialization, which becomes negligible in the long run.  

It is worth noting only a $29\%$ reduction in parameters in ResNeXt-29-Lr and an interesting $70\%$ increment in MobileNetV2-Lr, both of which use group convolutions \cite{krizhevsky2012imagenet, xie2017aggregated}. This behavior can be explained based on Eq. \ref{eq:argmin_alp}. The requirement for parameter reduction is adjusted as the effective input channels $c_{l}$ becomes $c_{l}/g_{l}$ where $g_{l}$ is the number of groups. In MobileNetV2-Lr, this requirement is violated, resulting an increment in parameters. It further shows a notable increment of $\times 166$ in FLOPs, since the matrix dimensions become very large in this architecture with a large number of filters per layer, which does not affect the cost of depth-wise separable convolutions \cite{chollet2017xception} much. However, this behavior can still be contained in the rank-reduced version MobileNetV2-LRr, which shows a $41\%$ reduction in parameters with $\times 8.5$ increment in FLOPs. 
\newcolumntype{L}[1]{>{\raggedright\arraybackslash}p{#1}}
\begin{table}[t]
	\small
	\begin{center}
		\resizebox{0.48\textwidth}{!}{
			\begin{tabular}{l P{14mm}P{12mm}P{12mm}P{13mm}P{10mm} }			
				\hline
				Model   & CIFAR-10 (\%) & SVHN \quad(\%) & ImageNet (\%) & Parameters (M) & FLOPs \quad(B) \\ \hline
				SqueezeNet \cite{hu2018squeeze}  & $83.8$ & $95.1$ & $71.4$ & $0.73$ & $0.069$ \\ 
				SqueezeNet-LRr (ours) & \footnotesize-1.5 & \footnotesize+0.2 & \footnotesize-0.9 & $0.49$ & $0.079$ \\ \hline
				VGG11 \cite{simonyan2014very}      & $90.4$  & $95.4$ &  $70.2$ & $9.23$ & $0.171$ \\ 
				VGG11-OctConv \cite{chen2019drop}  & \footnotesize-1.1& \footnotesize-0.5  & \footnotesize-0.7 &  $9.23$ & $0.077$ \\ 
				VGG11-Basis \cite{li2019learning}  & \footnotesize-0.1& \footnotesize+0.2  & \footnotesize-0.2 &  $^*1.99$ & $^*0.101$ \\
				VGG11-LRr (ours)  & \footnotesize-0.6& \footnotesize+0.0  & \footnotesize-0.3 &  $4.65$ & $0.350$ \\\hline
			\end{tabular}
		}\vspace{3mm}
		\caption{Comparison of LinearConv models with some methods in literature. We show that LinearConv can be applied in complement with other architectures proposed for parameter reduction, such as SqueezeNet \cite{hu2018squeeze}. However, our method which requires no change in the architecture or the hyperparameter settings is more meaningful to be compared against such designs with minimal changes. LinearConv models show comparable performance with parameter reduction. Here, the values denoted with $^*$ are the requirements in addition to the baseline values, as the Basis filters \cite{li2019learning} require already trained baseline weights for training. This means it has a cumulative complexity more than that of the baseline. In contrast, our method can be trained from scratch with the reported complexity.
		}
		\label{Table:comp}
		\vspace*{-1mm}
	\end{center}
\end{table}

We further compare LinearConv models with some methods in literature for network compression as presented  in Table \ref{Table:comp}. Our proposed method is intended for parameter reduction without having any change in the network architecture---in which Conv layers are replaced---or the hyperparameter settings. With such an implementation, our method is widely applicable to many existing architectures as a convenient plug-and-play module. Based on this design rule, we compare our work against the methods in literature with minimal changes in architecture \cite{chen2019drop, li2019learning}. However, we show that our method can also be complementary in architectures specifically designed for parameter reduction, such as SqueezeNet \cite{hu2018squeeze}. VGG11-OctConv \cite{chen2019drop} reduces computational requirement, having the same number of parameters, whereas VGG11-Basis \cite{li2019learning} reduces both the number of parameters and FLOPs. However, VGG11-Basis requires the weights of the respective baseline for training, as it tries to approximate weights of the baseline. Therefore, in practice, the actual number of parameters and the computational requirement for training are the \textit{cumulative} amounts including those of the baseline. In contrast, LinearConv models have no such requirement and can be trained from scratch with the presented complexity. SqueezeNet-LRr shows further parameter reduction compared to its baseline, verifying that LinearConv can be complementary to other architectures for parameter reduction. We further apply LinearConv in EfficientNet \cite{tan2019efficientnet}, where it behaves as in MobileNetV2 due to depth-wise separable convolutions.

\begin{figure}[t]
	\vspace{-3mm}
	\begin{subfigure}{.5\textwidth}
		\centering
		\hspace{-5mm}
		\includegraphics[width=.9\linewidth]{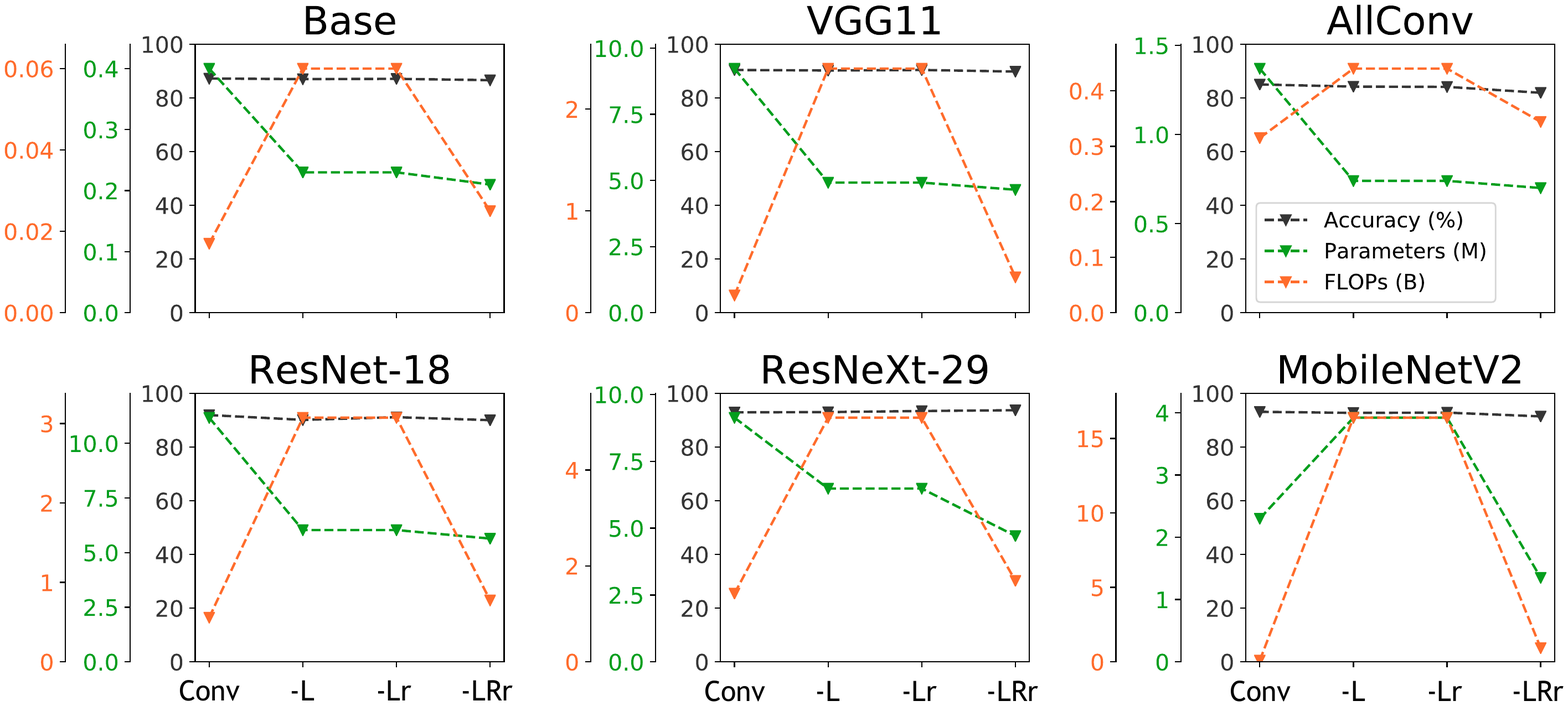}  
	\end{subfigure}
	\begin{subfigure}{.5\textwidth}
		\centering
		\vspace{-12mm}
		\includegraphics[width=.9\linewidth]{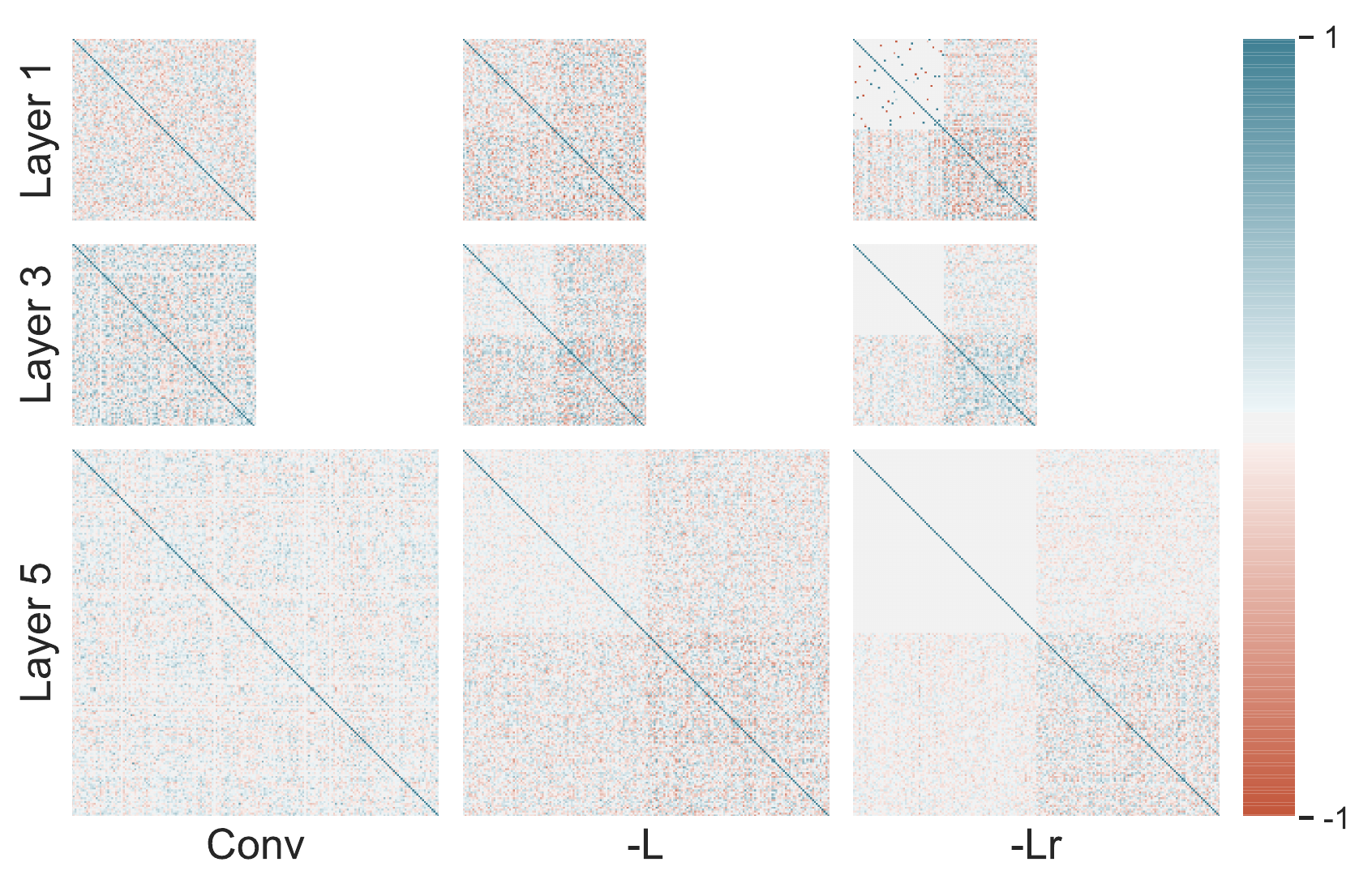}  
	\end{subfigure}
	\vspace{-2mm}
	\caption{(Top) Performance of LinearConv models on CIFAR-10. The regularized, rank-reduced version (-LRr) shows a performance similar to the baseline, with almost half the number of parameters and a similar computational requirement. 
		(Bottom) Correlation matrices corresponding to the weights in Conv, LinearConv (-L) and regularized LinearConv (-Lr). In LinearConv versions, the first and the second halves of each dimension in correlation matrices correspond to $\mathbf{W}^{p}_{l}$ and $\mathbf{W}^{s}_{l}$ respectively. When regularized, we can see that the primary filters become orthogonal (except for some artifacts in Layer 1). 
		\vspace{1mm}
	}
	\label{Fig.corr}
\end{figure}

Fig. \ref{Fig.corr} (Left) visualizes the performance graphs of Conv and LinearConv models in the CNN architectures that we selected. LinearConv models achieve a similar performance as the respective baselines with almost a $50\%$ reduction in parameters on average. The increment of computational requirement in LinearConv can be contained with the proposed rank-reduced implementation. Fig. \ref{Fig.corr} (Right) shows the correlation matrices corresponding to a few layers in Conv and LinearConv models. This visualizes the effect of regularization in the primary filters of LinearConv. The secondary filters generated as linear combinations show increased correlations as expected. We observe some artifacts in the correlation matrix which corresponds to the primary filters in Layer 1. This is due to a small non-zero regularization loss at the end of training.
\vspace{-2mm}

\section{Conclusion}
\label{se:conclusion}
In this work, we exploited the feature redundancy in CNNs to improve their efficiency. Based on the intuition of restricting the inherent redundancy and re-applying a sufficient amount to maintain performance, we proposed a novel layer: LinearConv, which can replace the conventional convolutional layers with a smaller number of parameters. We design LinearConv as a plug-and-play module to allow convenient adoption in existing CNNs with no change in the architecture or the hyperparameter settings. In combination with the proposed correlation-based regularization loss, LinearConv layers flexibly reproduce the redundancy in convolutional layers with almost a $50\%$ reduction in parameters on average, and the same computational requirement at inference. We evaluate its performance in common  architectures and show that it can function in complement with other methods in literature.  The control over feature correlation and redundancy opens-up room for improving the efficiency of CNNs. 
%

\pagebreak
\textbf{Acknowledgments:} The authors thank D. Samaras, C.U.S. Edussooriya, P. Dharmawansa, M. Pathegama and D. Tissera for helpful discussions. The authors acknowledge the computational resources received from the QBITS Lab and the Faculty of Information Technology, University of Moratuwa. K. Kahatapitiya was supported by the Senate Research Committee Grant no. SRC/LT/2016/04.

{\small
	\balance
	\bibliographystyle{ieee}
	\bibliography{egbib}
}

\pagebreak
\onecolumn
\section*{Appendix A:}

\setlength\extrarowheight{2pt}
\newcolumntype{C}[1]{>{\centering\arraybackslash}p{#1}}

{\setlength{\extrarowheight}{2pt}
	\begin{table}[hbt!]
		\small
		\begin{center}
			\resizebox{0.95\textwidth}{!}{
				\begin{tabular}{c|C{20mm}|C{20mm}|C{20mm}|C{25mm}|C{35mm}|c}
					\hline
					\vspace{10mm}Output & Base & VGG11 [46] & AllConv [48] & ResNet-18 [19] & ResNeXt-29 [56]  & MobileNetV2 [44] \vspace*{-10mm}\\
					\hline
					$32\times32$ & - & - & $\begin{array}{c}3\times3,\;96\\3\times3,\;96 \end{array}$  & $\begin{array}{c} 3\times3,\;64 \\ \left[\begin{array}{c} 3\times3,\;128\\3\times3,\;128 \end{array}\right]^2 \end{array}$& $\begin{array}{c} 1\times1,\;64 \\ \left[\begin{array}{c} 1\times1,\;128\\3\times3,\;128\\1\times1,\;256 \end{array};\;C=2\right]^3 \end{array}$ & $\begin{array}{c} 3\times3,\;32 \\ \left[\begin{array}{c} 1\times1,\;32\\3\times3,\;32\\1\times1,\;16 \end{array};\;C=32\right]^1\\ \left[\begin{array}{c} 1\times1,\;96\\3\times3,\;96\\1\times1,\;24 \end{array};\;C=96\right]^2 \end{array}$ \\ [48pt]
					\hline \rule{0pt}{6.5ex}  
					$16\times16$ & $3\times3,\;32$ & $3\times3,\;64$ & $\begin{array}{c} 3\times3,\;96 \\3\times3,\;192 \\3\times3,\;192 \end{array}$ & $\left[\begin{array}{c} 3\times3,\;128\\3\times3,\;128 \end{array}\right]^2$ & $\left[\begin{array}{c} 1\times1,\;256\\3\times3,\;256\\1\times1,\;512 \end{array};\;C=2\right]^3$ & $\left[\begin{array}{c} 1\times1,\;144\\3\times3,\;144\\1\times1,\;32 \end{array};\;C=144\right]^3$ \\ [18pt]
					\hline \rule{0pt}{12ex} 
					$8\times8$ & $3\times3,\;64$ & $\begin{array}{c} 3\times3,\;128 \\ 3\times3,\;256 \end{array}$ & 
					$\begin{array}{c}3\times3,\;192\\ 3\times3,\;192\\ 1\times1,\;192 \\1\times1,\;10\end{array}$ & $\left[\begin{array}{c} 3\times3,\;256\\3\times3,\;256 \end{array}\right]^2$ & $\left[\begin{array}{c} 1\times1,\;512\\3\times3,\;512\\1\times1,\;1024 \end{array};\;C=2\right]^3$ & $\begin{array}{c} \left[\begin{array}{c} 1\times1,\;196\\3\times3,\;196\\1\times1,\;64 \end{array};\;C=196\right]^4 \\ \left[\begin{array}{c} 1\times1,\;384\\3\times3,\;384\\1\times1,\;96 \end{array};\;C=384\right]^3\end{array}$ \\ [40pt]
					\hline \rule{0pt}{13.9ex} 
					$4\times4$ & $3\times3,\;128$ & $\begin{array}{c}3\times3,\;256 \\ 3\times3,\;512\end{array}$ & - & $\left[\begin{array}{c} 3\times3,\;512\\3\times3,\;512 \end{array}\right]^2$ & - & $\begin{array}{c} \left[\begin{array}{c} 1\times1,\;576\\3\times3,\;576\\1\times1,\;160 \end{array};\;C=576\right]^3 \\ \left[\begin{array}{c} 1\times1,\;960\\3\times3,\;960\\1\times1,\;320 \end{array};\;C=960\right]^1 \\1\times1,\;1280\end{array}$ \\
					\hline
					$2\times2$ & $3\times3,\;256$ & $\begin{array}{c}3\times3,\;512\\ 3\times3,\;512 \end{array}$ & - & - & - & - \\	
					\hline
					$1\times1$ & $\text{10-d \textit{fc}}$ & $\begin{array}{c} 3\times3,\;512 \\\text{10-d \textit{fc}} \end{array}$ & $\text{avg pool}$ &  $\begin{array}{c}\text{avg pool} \\ \text{10-d \textit{fc}}\end{array}$  & $\begin{array}{c}\text{avg pool} \\ \text{10-d \textit{fc}}\end{array}$& $\begin{array}{c}\text{avg pool} \\ \text{10-d \textit{fc}}\end{array}$ \\
					\hline 
				\end{tabular}
			}
			\caption{ Network architectures, in which we replace Conv with LinearConv in our experiments. [.]$^k$ represents $k$ repetitions of blocks having shortcut connections, and $C$ stands for the number of groups in group convolutions. Each Conv layer is represented as $h_{l}\times w_{l},f_{l}$. Output is the spatial dimension of feature maps $H_{l}\times W_{l}$.\vspace*{3mm}
			} 
			\label{Table:config}
			
		\end{center}
	\end{table}

\end{document}